\documentclass[runningheads]{llncs}
\usepackage[T1]{fontenc}
\usepackage{graphicx}
\usepackage{amsmath}
\usepackage{amsfonts}
\usepackage{subfigure}
\usepackage{caption}
\usepackage{subcaption}
\begin{document}
\title{Navigation and Exploration with Active Inference: from Biology to Industry}
%
%
\author{Daria de Tinguy\inst{1}\orcidID{0000-0003-1112-049X} \and
Tim Verbelen\inst{2}\orcidID{0000-0003-2731-7262} \and
Bart Dhoedt\inst{1}\orcidID{0000-0002-7271-7479}}
\authorrunning{de Tinguy D. et al.}
%
\institute{Ghent university, Ghent, Belgium
\email{daria.detinguy at ghent.be}\\ \and
Verses, Los Angeles, California, USA}
\maketitle              
\begin{abstract}
By building and updating internal cognitive maps, animals exhibit extraordinary navigation abilities in complex, dynamic environments. Inspired by these biological mechanisms, we present a real-time robotic navigation system grounded in the Active Inference Framework (AIF). Our model incrementally constructs a topological map, infers the agent’s location, and plans actions by minimising expected uncertainty and fulfilling perceptual goals without any prior training. Integrated into the ROS2 ecosystem, we validate its adaptability and efficiency across both 2D and 3D environments (simulated and real-world), demonstrating competitive performance with traditional and state-of-the-art exploration approaches while offering a biologically inspired navigation approach.

\end{abstract}
\section{Introduction}

Animals exhibit remarkable navigation capabilities that allow them to thrive in complex and unpredictable environments. From migratory birds flying across continents~\cite{migration}, to rodents learning intricate mazes~\cite{mice_in_labyrith} and humans navigating subways~\cite{humans-hierarchic-plan-subway}, biological agents demonstrate an ability to explore, localise, and plan. These capabilities rely on internal models of the world, enabling organisms to imagine trajectories and evaluate outcomes on the fly. Such abilities far exceed those of most autonomous systems, particularly in real-world settings where pre-mapped routes or supervised learning often fail to adapt to dynamic changes~\cite{orbslam3,AGVs,NNslam,toward_int_nav}.


Neuroscience has suggested that animals navigate using structured internal representations (cognitive maps) combining spatial, temporal, and relational information~\cite{Human_rodent_spatial_rep,humans-cognitive-map,humans-hierarchic-plan-subway}. Such models are formed and refined through experience, enabling organisms to infer location, predict outcomes, and choose actions with minimal supervision. Translating these insights to robotics could enable adaptive, data-efficient, and robust behaviour~\cite{learn_to_nav,exp_learning_survey}.


The Active Inference Framework (AIF) offers a principled and biologically grounded approach to modelling such behaviour, proposing an unifying framework for this endeavour. Rooted in Bayesian inference and predictive coding, AIF treats perception, learning, and action as components of a single generative process: agents minimise expected free energy by updating beliefs about the world and selecting actions to reduce uncertainty or satisfy preferences~\cite{life_friston,world_model_and_inference}.
While conceptually appealing and biologically plausible, AIF has seen limited application in embodied robotics, with only a few recent works exploring its real-world deployment in the context of navigation~\cite{GSLAM}.


In this paper, we present a bio-inspired navigation system that applies AIF to real-world spatial reasoning, localisation, and planning. Our agent builds and updates a topological cognitive map on the fly, using it to infer its position and state, evaluate hypotheses about unvisited areas, and choose actions that minimise expected surprise. It requires no pre-training, is robust to sensor drift and environmental change, and adapts continuously.


We validate our approach in a variety of settings, from simple 2D mazes to large-scale 3D environments (using ROS2~\cite{ros2}) with realistic dynamics and sensing. These include structured and unstructured indoor scenarios such as warehouses and a real-world maze. Our results show interpretable and adaptive behaviour, supporting efficient planning and generalisation without heavy data training, moving robotic autonomy closer to the flexibility of natural navigation.

The paper is structured as follows: We begin by reviewing navigation strategies that support exploration under Active Inference. We then present our method for topological mapping, belief-based localisation, and goal-directed planning, along with its application in 2D mazes. The system architecture and its ROS2 implementation are described next, followed by a comprehensive evaluation of its exploration efficiency in 3D environments (simulated and real-world). We conclude by discussing current limitations and potential extensions.

\section{Related Work}



Navigation requires integration of localisation and mapping, decision-making (where should I go), and motion planning (how should I move). Those are typically separated in classical navigation pipelines, often focusing on a subset of navigation and relying on hand-tuned rules or pre-training. Hand-tuned rule models often have limited adaptability in dynamic settings as well as potentially computationally intensive or poorly scalable to large environments~\cite{orbslam3,AGVs}, while learning-based methods require extensive environment-specific data and struggle to generalise~\cite{exp_learning_survey}. 

Recent work combines mapping and planning via probabilistic or topological models~\cite{ETPNav,gaussian_process_nav}. They are lightweight models, scalable, and well-suited for reasoning and planning in complex environments. However, these models often depend on costly computations or fail to generalise in large, unstructured and dynamic environments. Similarly, self-supervised and dataset-driven methods like BYOL-Explore~\cite{BYOL} or ViKiNG~\cite{viking} show strong performance in, respectively, exploration or goal-reaching; however, they require an important training phase in similar environments, and only fulfil a given task.

Conversely, Active Inference offers a unified, biologically inspired alternative that treats navigation as inference, avoiding explicit reward functions. Rooted in the idea that agents minimise surprise through belief updating, AIF enables continuous adaptation by integrating perception, localisation, and action selection~\cite{AIF_book}.

Initial works demonstrated how AIF could support emergent animal behaviours like exploration and goal-reaching~\cite{Human_rodent_spatial_rep} in simplified 2D settings~\cite{nav_aif,curiosity_exploitative,weird_HAIF}. More recent studies introduced cognitive maps and structure learning~\cite{Toon_integration}, or proposed hierarchical models for efficient spatio-temporal planning~\cite{ours_hierarchy}. While these contributions highlight the flexibility of AIF, they remain limited to low-dimensional, 2D environments and often rely on computationally expensive policy search.

Notably, G-SLAM~\cite{GSLAM} demonstrated real-world AIF deployment via unified mapping and control, but required pre-trained generative models and lacked compatibility with robotics stacks.

Our work builds on these ideas to demonstrate how AIF can be applied in real-world navigation scenarios without requiring prior training or rigid pipelines.
We propose a model for robotic navigation that efficiently explores fully unknown environments using modular components (with ROS2) and flexible sensory input to achieve joint mapping, localisation, and decision-making.

\section{Inferring Motion to Foretell Model Growth}

When navigating an environment, an AIF agent continuously refines its internal model of the world by updating transition probabilities (the likelihood of moving between states) and observation likelihoods (the probability of perceiving certain features given a state). These updates are driven by the agent's actions and resulting sensory inputs, gradually aligning the model's predictions with actual observations. This alignment allows the agent to better anticipate outcomes and select actions, minimising Expected Free Energy (EFE), to optimise its navigation strategy.

\begin{figure}[!htb]
    \centering
    \includegraphics[width=6cm]{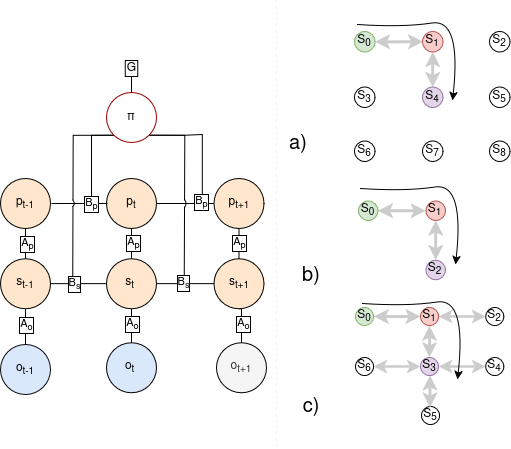}
    \caption{Left: Factor graph of the POMDP generative model, showing transitions from past to future (up to time $t+1$). Known observations (blue) inform current latent states. Future actions follow policy $\pi$, influencing inferred positions and states (orange), and generating predictions of future observations (grey). The agent's position $p_t$ is determined by $p_{t-1}$ and the selected policy, while the latent state $s_t$ is inferred from $o_t$, $p_t$, and $s_{t-1}$. Transitions are parametrised by $B$ matrices, and $A$ matrices encode the likelihood of observations given latent states.
Right (a–c): Three ways of structuring the world, progressing from the most common (a) with a given static world dimension, to (b) a growing state learning given a new observation, to (c) the structure learned by our proposed model given expected motions.}
    \label{img:pomdp}
    \vspace{-5mm}
\end{figure}

A key limitation in most advanced models is that they 1) have static state dimensions (illustrated Figure~\ref{img:pomdp} a) and presented in ~\cite{weird_HAIF,Toon_integration}) and/or 2) often expand their internal state space only after encountering new observations, as pictured in Figure~\ref{img:pomdp} b) and presented in~\cite{surpervised_struct_learning,bayesian_model_reduc}. For instance, in room-structured mazes, new states are added only when the agent physically enters a new room. However, this strategy causes the agent to forget unexplored possibilities, such as unvisited rooms observed indirectly (e.g., through visible doors), and thus cannot use this information to predict or plan future exploration effectively. As a result, policies evaluated through EFE become short-sighted, focusing only on directly experienced transitions and ignoring potentially informative unvisited areas.


To overcome this, our model expands its internal topological map based not only on actual observations but also on predicted states (as presented in Figure~\ref{img:pomdp} c)). By considering hypothetical transitions (such as unvisited rooms behind seen doors), the agent maintains connectivity across the environment and can revisit these paths later. This allows the agent to retain awareness of unexplored options and plan accordingly. This strategy significantly improves exploration efficiency, enabling more comprehensive and strategic coverage of the environment with less redundancy and delay.


The core of our model relies on the generative model presented in Figure~\ref{img:pomdp}, where we separately model the inference of the location (state $s$) from the estimated position $p$ as two distinct latent variables.  
$p$ encodes the believed position of the agent considering motion, past position and past state confidence, while $s$ is the state representing the localisation of the agent, consisting of an observation and a position. $p$ is deduced from the previous action $a_{t-1}$, pose $p_{t-1}$ and state $s_{t-1}$. While $s$ is inferred from the current observation $o$ (from which we expect to deduce the presence of obstacles), position $p$ and previous state $s_{t-1}$. Inferring those two variables improves the robustness of the system to kidnapping and ambiguous situations (where an observation changed in a past location). 
\vspace{-5mm}

This model results in the approximate posterior presented in Equation~\eqref{eq2}
\begin{equation} 
Q(\tilde{s},\tilde{p}| \tilde{o}, \tilde{a}) = Q(s_0, p_0| o_0) \prod_{t=1}^\tau
Q(s_t, p_t| s_{t-1},p_{t-1}, a_{t-1}, o_t )  
\label{eq2}
\end{equation} 

Our model’s key component is its ability to infer both position and latent state jointly. This enables it to remain robust under visual ambiguity (e.g., perceptual aliasing or kidnapping) while also supporting self-expansion by hypothesising unexplored parts of the environment.

Expanding the agent's model is also governed by Free Energy minimisation, i.e. by comparing whether an expanded model better explains current or expected observations than our current model at hand. Concretely, given a current model $P$, and an alternative, expanded model $\tilde{P}$, the Free Energy difference can be written as~\ref{eq:PvsP}~\cite{AIF_book}:

\begin{equation}
\Delta F = F[\tilde{P}(\theta)] - F[P(\theta)] = \ln \mathbb{E}_{Q(\theta)} \left[ \frac{P(\theta)}{\tilde{P}(\theta)} \right]
\label{eq:PvsP}
\end{equation}

 If the free energy of an expanded model is lower than that of the current model, the agent updates its internal structure to incorporate the newly encountered (or predicted) information, effectively increasing the model's dimension. This involves predicting motions leading to previously uncharted states through the Expected Free Energy (EFE) of policies, where the benefit of updating $A_p$ is evaluated. 

The generative model includes a state transition matrix $B_s$ and an observation likelihood matrix $A_o$. In addition, it features a position likelihood matrix $A_p$, linking each state to a possible pose, and a transition model $B_p$, which differs from the standard matrix form. $B_p$ is implemented as a list that tracks imagined positions. The agent infers its next pose by updating the previous position based on the intended action $a$ and the expected collision outcome $P(c)$, a binary variable (1 if we expect an obstacle between two poses or 0 otherwise). The EFE of a policy $\pi$ is defined in Equation~\ref{eq:efe_wt_c}. The learning term of this equation quantifies how much we learn about position likelihood in light of possible obstacles, while the inference term evaluates the next state $s_{t+1}$ and position $p_{t+1}$ given an expected collision $c_{t+1}$. This mechanism assumes that the agent can detect or identify obstacles via its observations $o$.


\begin{equation}
\begin{aligned}
G(\pi) &= \mathbb{E}_{Q_{\pi}} [\log Q(s_{t+1},p_{t+1}, A_p | \pi) - \log Q(s_{t+1},p_{t+1}, A_p | c_{t+1}, \pi) - \log P(c_{t+1})] \\
&= -\underbrace{\mathbb{E}_{Q_{\pi}} [\log Q(A_p |s_{t+1}, p_{t+1}, c_{t+1}, \pi) - \log Q(A_p |s_{t+1}, p_{t+1}, \pi)]}_\text{expected information gain (learning)}\\
&\quad -\underbrace{\mathbb{E}_{Q_{\pi}} [ \log Q(s_{t+1},p_{t+1} | c_{t+1}, \pi) - \log Q(s_{t+1},p_{t+1} | \pi) ]}_\text{expected information gain (inference)} \\
&\quad -\underbrace{\mathbb{E}_{Q_{\pi}} [ \log P(c_{t+1}) ]}_\text{expected collision}
\end{aligned}
\label{eq:efe_wt_c}
\end{equation}

Policies $\pi$ may lead the agent toward locations that are not yet represented in the generative model. To decide whether to expand the model to include such locations, we evaluate Equation~\ref{eq:efe_over_param}, which computes the EFE of the prior over the position likelihood parameters $A_p$. A high expected information gain from exploring a particular direction suggests that the model should grow to accommodate a new pose. However, if a collision $c$ is likely, this suppresses the probability of creating a new position at that location.


\begin{equation}
\begin{aligned}
P(A_p) &= \sigma(-G) \\
G(A_p) &= \mathbb{E}_{Q_{A_p}} [\log P(p,s |A_p) - \log P(p, s |c, A_p) - \log P(c)] \\
&= -\underbrace{\mathbb{E}_{Q_{A_p}} [\log P(p, s|c, A_p) - \log P(s |p,A_p) - \log P(p|A_p)]}_\text{expected information gain (expanding)} \\
&\quad  
- \underbrace{\mathbb{E}_{Q_{A_p}} [\log P(c)]}_\text{expected collision}
\end{aligned}
\label{eq:efe_over_param}
\end{equation}

If adding parameters to $A_p$ reduces EFE compared to the current model, then the model expands its parameter space to include a new position in $A_p$ and consequently $B_p$. This entails creating a new state associated with that position, which increases the dimensionality of all model components. The updated observation model $A_o$ assigns uniform probabilities to newly added states, reflecting the uncertainty about unvisited states. The transition model $B_s$ forms or updates transition probabilities between existing and newly created states, considering how much certainty we have about the new state. Details of this process are further discussed in Section~\ref{section:trs}.

To ensure robustness to kidnapping (sudden displacements) and observational ambiguity, the model relies on joint confidence in the inferred state and pose, conditioned on the previous state, pose, and action. If our current observation diverges from expectations by a given threshold but the state and pose transitions remain coherent, the agent updates its observation likelihood while maintaining confidence in its position. Conversely, if both the observation and the inferred transitions are surprising, the model reduces confidence in its current pose, alerting the model to a possible kidnapping. In such cases, the model halts further updates until a sequence of consistent observations restores confidence in the agent's location (confidence in a state reaches the given threshold), at which point the model updating process can resume.

\section{Exploration and Goal Reaching in a Tolman Maze}
\begin{figure}[!htb]
    \centering
    \includegraphics[width=7cm]{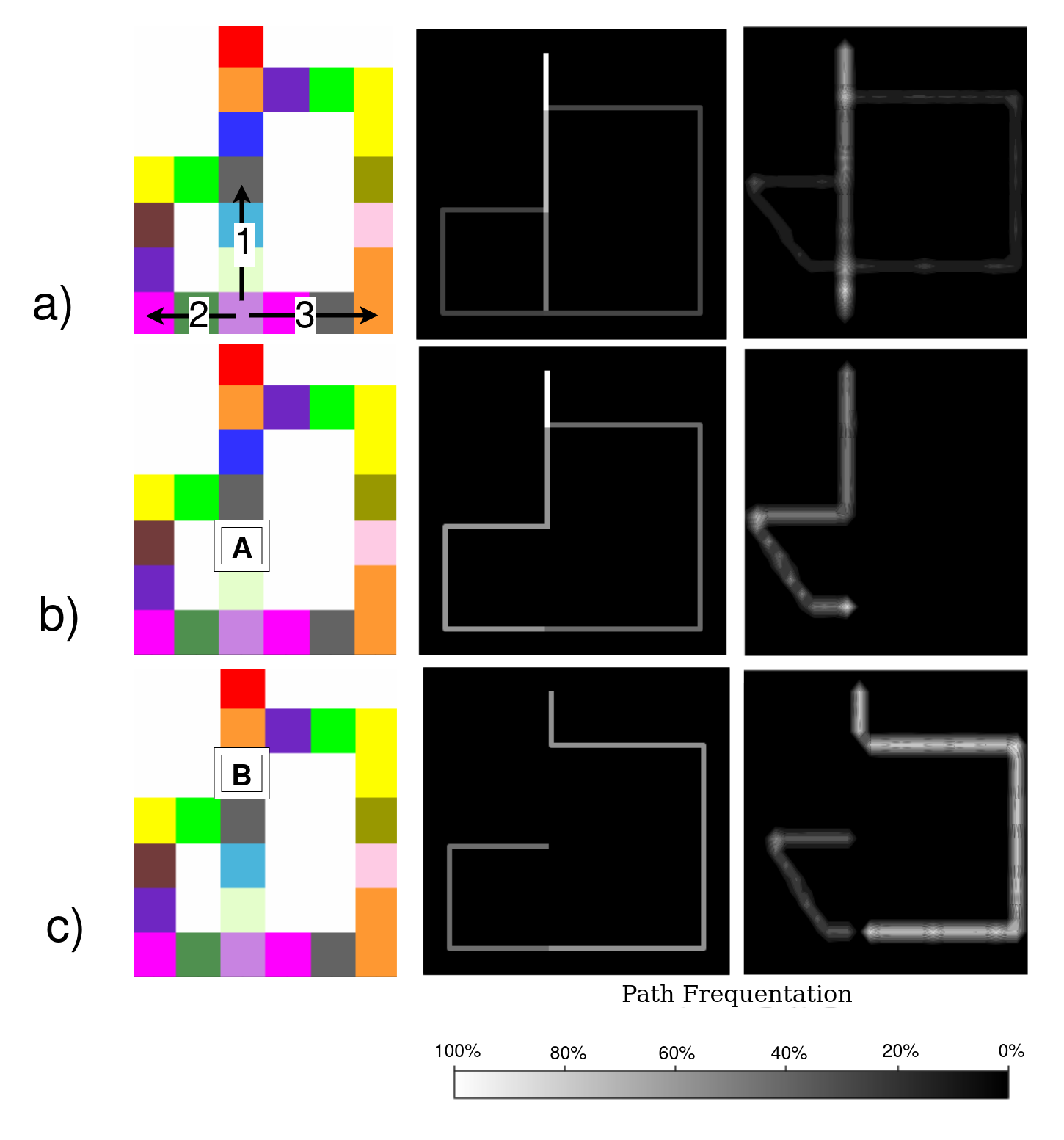}
    \caption{Our results (second column) compared to L.-E. Martinet \& al's (third column)~\cite{tolman_maze2_redone}. In our study, the agent's flow paths towards the objective (top of the map) are shown, with re-planning occurring when the desired path is blocked. The varying colour gradient of the lines indicates the frequency of selection for each path over all agents. A sequence is read horizontally, a) is the maze without obstacles, and b) and c) illustrate obstacles at points A and B, respectively. The occupancy grid maps demonstrate the learning of maze topology by simulated agents, initially without obstacles, showing a significant preference for Route 1. When a block is introduced at point A, the animals predominantly choose Route 2. With an obstacle placed at point B, the animals mainly opt for Route 3.}
    \label{img:tolman_maze}
    \vspace{-4mm}
\end{figure}

Our model's navigation strategy is evaluated in a dynamic maze environment inspired by the second maze of Tolman's experiment~\cite{Tolman_maze}, shown in Figure~\ref{img:tolman_maze} second column. 
The original experiment aimed to show evidence of the development of a flexible 'cognitive map' of the maze during exploration and that such mental representation guided the rat's behaviour in the reward sessions. Calling this mental map reorganisation "insight", defined later as "the solution of a problem by the sudden adaptive reorganization of experience"~\cite{Insight_def}. 

Agents begin at a designated start point (bottom of the maze) and must locate a reward placed at the opposite side. Three possible paths connect the start and goal, with Route 1 being the shortest and Route 3 the longest. Two critical junctions, A and B, can be blocked to restrict access to Routes 1 and 2, respectively.

Agents start with no prior knowledge of the maze structure or sensory cues, but are guided by a preference for red tiles. A utility weight of 2 biases the agent toward preference-seeking over pure exploration. Learning is cumulative across all conditions, allowing agents to build and update their knowledge acquired in earlier runs.

We ran ten agents under three sequential conditions: 
\begin{itemize}
    \item no obstacles, Figure~\ref{img:tolman_maze} a)
    \item a blockage at A, Figure~\ref{img:tolman_maze} b)
    \item a blockage at B, Figure~\ref{img:tolman_maze} c)
\end{itemize}
Each condition involves 12 sequential runs per agent. Every 20 time steps, the agent is 'kidnapped' and repositioned at the start without notice, requiring it to correct beliefs and infer its new location using sensory inputs.

Figure~\ref{img:tolman_maze} shows the frequency of path selections between conditions of our model (second column), compared to the results from~\cite{tolman_maze2_redone} (third column), which uses 100 agents. Unlike those agents or rats in the original experiment, ours cannot detect obstacles from afar; it must reach adjacent rooms to perceive blockages. Cells correspond to discrete rooms, with white blocks indicating obstructions. The agent’s route decisions thus emerge from internal belief updating given direct perception of new obstacles, showing how the agent tends to choose the most efficient path depending on the situation. Full scenario details and conclusions are provided in~\cite{ours_model}. 

The observed adaptability reflects two core mechanisms: the agent’s capacity to simulate action outcomes up to 14 steps ahead and the plasticity of its internal map (its ability to flexibly revise beliefs based on new evidence and adapt to a kidnapping situation). These traits highlight our model's potential for efficient, insight-driven navigation under uncertainty in simple 2D environments.

\section{Toward realistic settings}
\label{section:trs}
In this section, we are bridging the gap between our principled model working in low-dimensional settings and our model able to cope with the messiness of real-world environments.

To assess the feasibility of deploying our approach under realistic conditions, we implement it on a physical robot operating in previously unseen realistic indoor spaces. This setting challenges the model with sensor noise and non-uniform geometry, aspects often simplified in 2D environments.

Our system is deployed with ROS2 and is modular by design. The AIF-based planner is module, sensor and platform-agnostic and can interact with traditional perception modules and motion planning systems as well as diverse robots (Turtlebot, Turtlebot3~\cite{turtlebots} and RosbotXL~\cite{rosbotxl}, the used robots and sensors are defined in Appendix~\ref{app:robots}). Each component can be independently replaced or refined without disrupting the rest of the pipeline. Future work could incorporate more sophisticated perceptual pipelines (e.g., semantic SLAM or learned embeddings), without requiring changes to the generative model. 

\begin{figure}[!htb]
    \centering
    \includegraphics[width=0.45\linewidth]{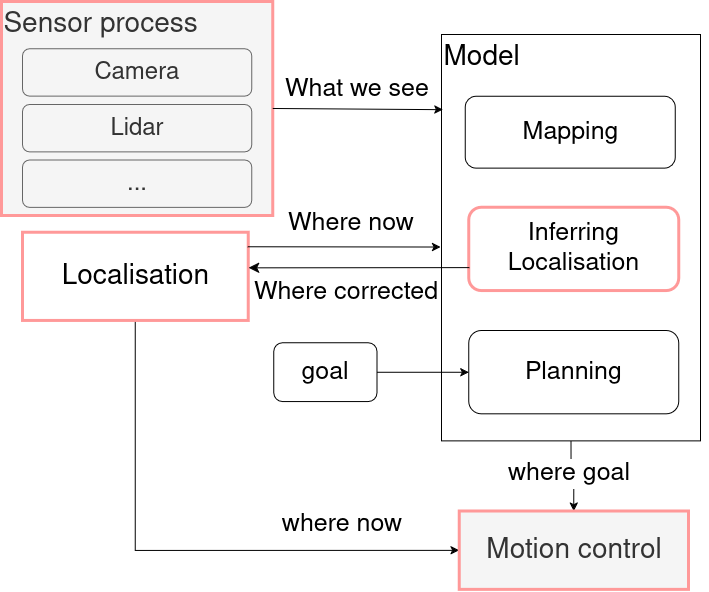}
    \caption{Overview of the system architecture. Modules interact through belief propagation, Inferring and planning (localisation, mapping and action selection) rely on the Active Inference framework. The perceptual and motion planning still use traditional approaches. Believed odometry takes precedence over sensor odometry. Preferences are expected from the user if we want to reach a target observation. Red contours highlight newly added or modified modules for Real-world navigation.}
    \label{fig:archi}
    \vspace{-4mm}
\end{figure}

The complete architecture of the navigation system is illustrated in Figure~\ref{fig:archi}. It is made of four modular components: (i) a generative model that performs mapping, inferring and planning under AIF framework, (ii) an odometry inference module (localisation) that estimates the state of the agent based on belief rather than direct sensor readings, (iii) a sensor processing stream (currently using panoramic visual input) and (iv) a motion control module responsible for the execution of selected actions. Modules newly added or modified from the 2D experiment to the real-world experiments are highlighted by red contours.

Critically, we do not rely on sensory estimated pose but on the agent's inferred pose, aligning with the epistemic stance of AIF, where belief takes precedence over sensory measurements. As a result, in the face of occlusion or localisation drift causing metric misalignment with the ground truth, the agent remains functionally coherent, navigating based on its internal generative model rather than needing external corrections.

Motion planning is handled by external controllers (e.g., Nav2~\cite{nav2} or potential field planners~\cite{potential_field}) that receive the position of the target state as input. 
However, goals are not specified as fixed coordinates but rather as probabilistic regions (Gaussian distributions) corresponding to expected poses and associated sensory observations. The agent determines it has reached a goal not through position alone, but when its actual sensory input aligns with its predicted perceptual state.

In deployment, we observe the system's ability to autonomously construct and extend its internal model while navigating through 3D spatial layouts. As exploration unfolds, the agent updates its representation to accommodate newly encountered scenes, refining its model via action-perception cycles. When getting to the next desired position, the agent captures a panoramic RGB view (taken with the camera and a rotation of the robot) and the latent cognitive states are updated via Structural Similarity Metrics (SSIM) between incoming views and stored experiences. When a discrepancy arises, it updates its internal model: either refining its observation model or revising its transition model, depending on its confidence in its current state estimate.

We use a Lidar to consider collisions as our camera does not provide accurate depth measurement; however, in our model, we still consider this as a visual observation $o$ and do not divide the information into separate observations. This results in a Transition matrix $B_s$ being updated according to the situation. We use a Dirichlet pseudo-count mechanism defined in Equation~\ref{eq:B_up} with learning rates ($\lambda$). The learning rates are determined by whether we were physically blocked toward an objective or we imagine being able to reach (or not) a position according to our sensors. The diverse situations are presented in Table~\ref{tab:tran_lr}. This continual adaptation of the state transitions according to the situation allows the agent to rapidly reconfigure its internal map to new or undetected obstacles (or removed obstacles) and avoid blocked regions, even several steps away, up to a user-set preferred distance, usually corresponding to the sensor limit. This increases the adaptability of the model to situations and better fine-tunes its internal map to the reality of the environment.

\begin{equation}
    B_\pi = B_\pi + Q(s_t|s_{t-1}, \pi) Q(s_{t-1}) * B_\pi  * \lambda
    \label{eq:B_up}
    \vspace{-2mm}
\end{equation}

\begin{table}[t!]
\vspace{-3mm}
    \centering
    \caption{Transition learning rate ($\lambda$) depending on the situation}
    \begin{tabular}{l|c|c|c|c}
        \textbf{Transitions} & \textbf{Possible} & \textbf{Impossible} & \shortstack{\textbf{Predicted} \\ \textbf{Possible}} & \shortstack{\textbf{Predicted} \\ \textbf{Impossible}} \\
        \hline
        Forward  & 7  & -7 & 5 & -5 \\
        \hline
        Reverse  & 5  & -5 & 3 & -3 \\
    \end{tabular}
    \vspace{-5mm}
    \label{tab:tran_lr}
\end{table}
\vspace{-2mm}
\section{Exploration in realistic environments}
\vspace{-3mm}
\begin{figure}[htb!]
\vspace{-5mm}
\centering    
\subfigure[280m$^2$ simulated warehouse]{\label{fig:map_big_ware}\includegraphics[width=.35\linewidth]{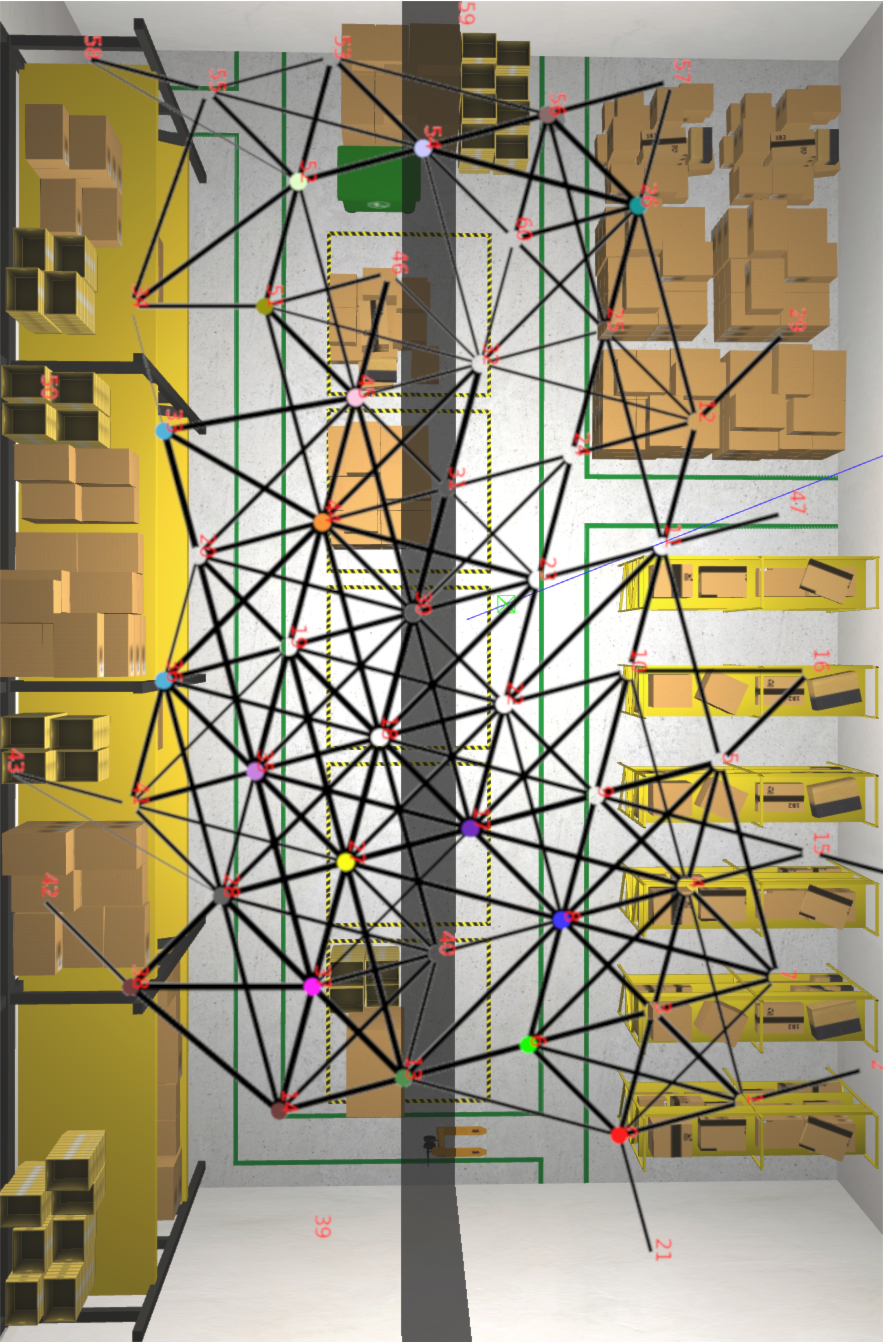}}
\subfigure[5m$^2$ of real environment]{\label{fig:map_maze_rw}\includegraphics[width=.35\linewidth]{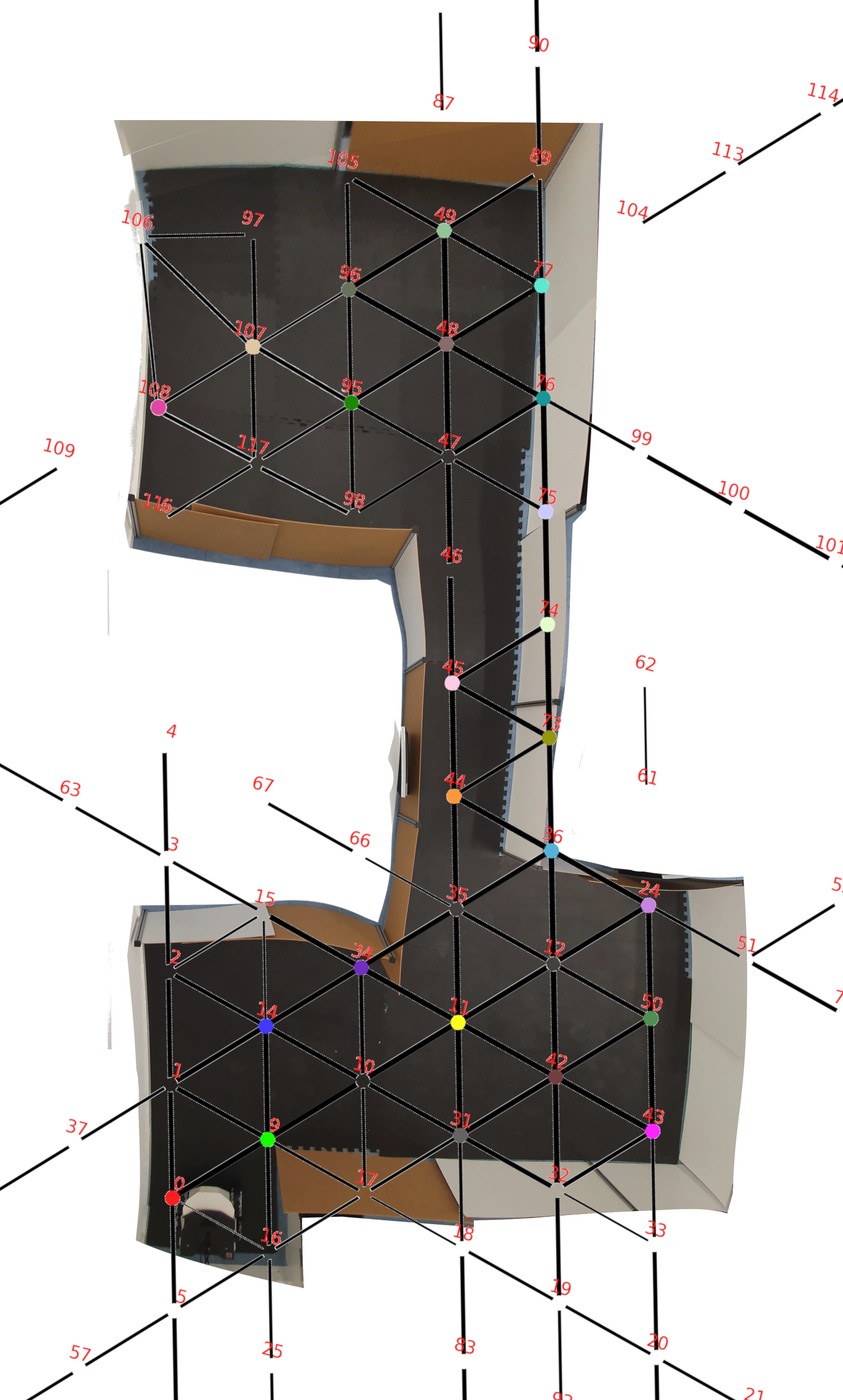}}
\caption{Final map of exploration in a) Amazon simulated warehouse, b) a real-world environment. Coloured points signify visited locations, where the same colour attributions mean the same observation. The thickness of the lines depicts the agent's believed probability of transitioning between two states given an action.}
\label{img:maps}
\vspace{-5mm}
\end{figure}

We evaluated our model in both simulated and real-world settings. The simulated environment, built in Gazebo~\cite{gazebo}, consists of a 280m$^2$ warehouse-like layout~\cite{warehouse}, shown in Figure~\ref{fig:map_big_ware}, while the real-world test was conducted in a 5m$^2$ controlled maze in Figure~\ref{fig:map_maze_rw}. In both cases, the agent incrementally builds its topological map using a user-defined spatial resolution, with a minimum node spacing of approximately 2m in simulation and 0.5m in the real environment. This minimum spacing represents the radius of influence of a state, and this adaptive granularity enables the agent to scale its representation to match the environmental complexity.
 
The agent's action space consists of 13 discrete actions: 12 evenly spaced orientations across 360°, plus a "stay" action. However, when expanding its topological map, a state can generate transitions up to a maximum of six adjacent states, rather than creating a new node for every possible direction. This constraint is imposed to prevent an overly dense graph structure and maintain a manageable level of connectivity. By limiting the number of newly created states per location, the agent ensures that only the most promising directions are used to expand the map. This approach preserves the flexibility to connect to more distant or informative states later, particularly when no obstacles are present, thereby maintaining a sparse yet navigable topological representation of the environment. Both the number of imagined transitions and available actions can be tuned by the user, allowing for control over planning granularity and computational cost.

Each visited location is associated with a sensory observation. According to the agent's observation model, locations sharing the same dot colour in the map visualisation indicate perceptual similarity. The thickness of the edges between the nodes represents the inferred likelihood of successful transitions, and thicker edges denote a greater confidence in the navigability between states. In the real environment, the lidar sometimes hallucinates free space due to incorrect reflections, and erased connections are visible when the agent attempts to reach those points. 
\begin{figure}[!htb]
    \centering
    \vspace{-7mm}\includegraphics[width=0.65\linewidth]{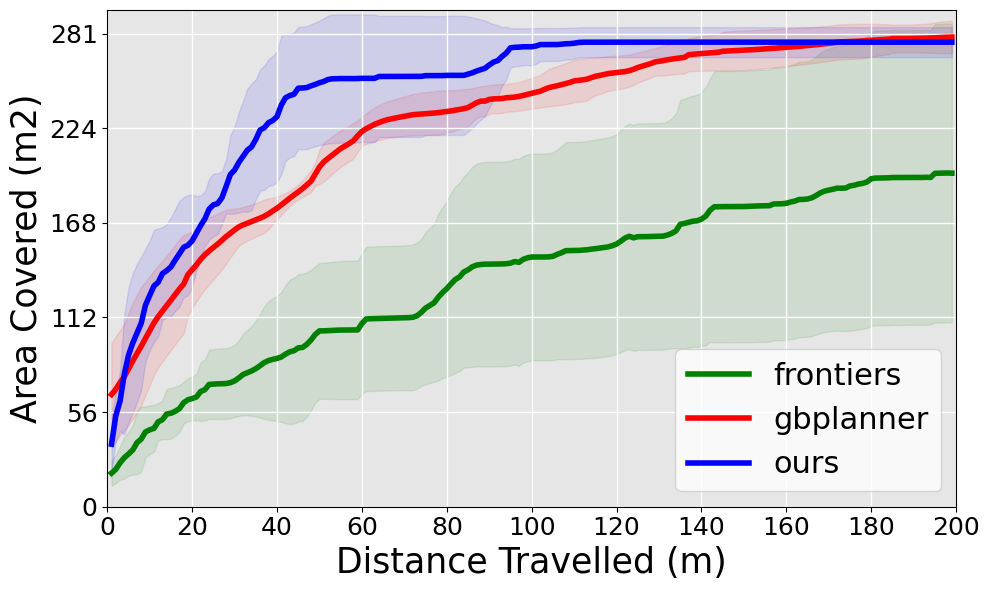}
    \caption{Lidar Coverage of a 280m$^2$ warehouse over the distance travelled (m) with our model, Frontiers and Gbplanner over five runs each.}
    \label{fig:int_map_nav}
    \vspace{-5mm}
\end{figure}

To evaluate exploration performance, we compared our model against Frontiers~\cite{frontiers}, which is a classical exploration algorithm aiming for unexplored areas and Gbplanner~\cite{gbplanner}. Gbplanner uses a metric map~\cite{Voxblox} to form its topological map, then used for exploration. Gbplanner is an improved version of the 2021 DARPA's winner~\cite{Darpa_winners}. While being an AI method, our model is not suitable for comparing our navigation to deep-learning methods due to the absence of pre-training.  

Figure~\ref{fig:int_map_nav} displays the area observed by the robot (considering the Lidar) over its travelled distance, averaged over five runs per model, initiating at different starting points in the warehouse. Time-based metrics were avoided due to variability in simulated time within the Gazebo environment.

Overall, our Active Inference framework outperforms or matches the performance of more traditional approaches (Frontiers) and state-of-the-art exploration systems (Gbplanner), with an efficient goal-oriented exploration (goals being the next state to reach). Frontiers falls behind in exploration efficiency as it lacks optimised navigation and requires multiple passes over the same areas. Especially when some small, unreachable areas have unexplored zones without a clear frontier. Gbplanner has been built to be robust in underground passages rather than optimising navigation in large open spaces.
Additional experiments validating the dynamic adaptability of our model can be found in appendix~\ref{app:obstacles}.

These experiments validate the applicability of our approach beyond simulation and into realistic settings, laying a foundation for more flexible, autonomous navigation in the world based on a biologically plausible strategy.

\vspace{-3mm}
\section{Conclusion}

We have illustrated how a biologically inspired navigation system grounded in the Active Inference Framework (AIF) could be used to navigate in an unknown environment without pre-training in 2D and 3D environments (simulated and real). By unifying probabilistic localisation, topological mapping, and belief-driven planning into a single generative model, our approach offers an alternative to traditional navigation pipelines that often rely on rigid assumptions, pre-training, or heavy computational overhead. Our model supports continuous learning and decision-making under uncertainty, leveraging expected free energy minimisation to guide exploration and goal-directed behaviour. The model shows promise in competing with traditional exploration strategies such as Frontiers~\cite{frontiers} or Gbplanner~\cite{gbplanner} in exploration strategy efficiency. 
In addition, our modular architecture ensures compatibility with standard robotics stacks (ROS2), allowing for straightforward integration with existing perception and control systems. 

While promising, our current model still faces limitations related to perception analysis. Future work will focus on precisely determining the limits of the current model in real-world situations (including computational load analysis) as well as enhancing perceptual inference (e.g., via semantic representations or learnt embeddings), extending hierarchical planning capacities, and improving computational efficiency for broader deployment. Ultimately, this work moves a step closer to biologically plausible and practically capable robotic navigation in open-ended, real-world scenarios.

\section*{Acknowledgements}

This research received funding from the Flemish Government (AI Research Program) under the “Onder-zoeksprogramma Artificiële Intelligentie (AI) Vlaanderen” programme and the Inter-university Microelectronics Centre (IMEC).

%
%
%
%
\bibliographystyle{splncs04}
\bibliography{main.bib}

\section*{Appendix}

\subsection*{Robots}
\label{app:robots}

Our system is robot-agnostic; however, we have to adapt the sensor pipeline to the specific sensors used. 
In simulation, we used a TurtleBot3 Waffle with a Pi camera and a 360-degree lidar with a 12 m range. 
In the real environment, several robots have been used; at first, we used the Turtlebot with a forward lidar of 240 degrees of 12 m range and a camera RealSense D435; the Turtlebot4 with a 360-degree lidar of 8 m range and a camera OAK-D-Pro. Due to uncorrected drift, the resulting maps did not allow a clear superposition with the layout of the environment; thus, we used a RosbotXL with a 360-degree lidar of 18m range and a 360-degree camera to better show the results of the navigation. However, while the Lidar range is 18m, the agent only considers the Lidar up to 8 consecutive nodes to create or update new transitions to be as reliable as possible.

\begin{figure}[htb!]
\vspace{-6mm}
\centering    
\subfigure[Turtlebot3: Waffle]{\label{fig:waffle}\includegraphics[width=.4\linewidth]{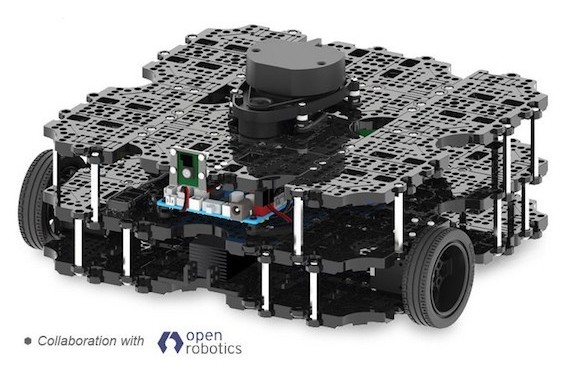}}
\subfigure[Husarion: RosbotXL]{\label{fig:rosbotxl}\includegraphics[width=.4\linewidth]{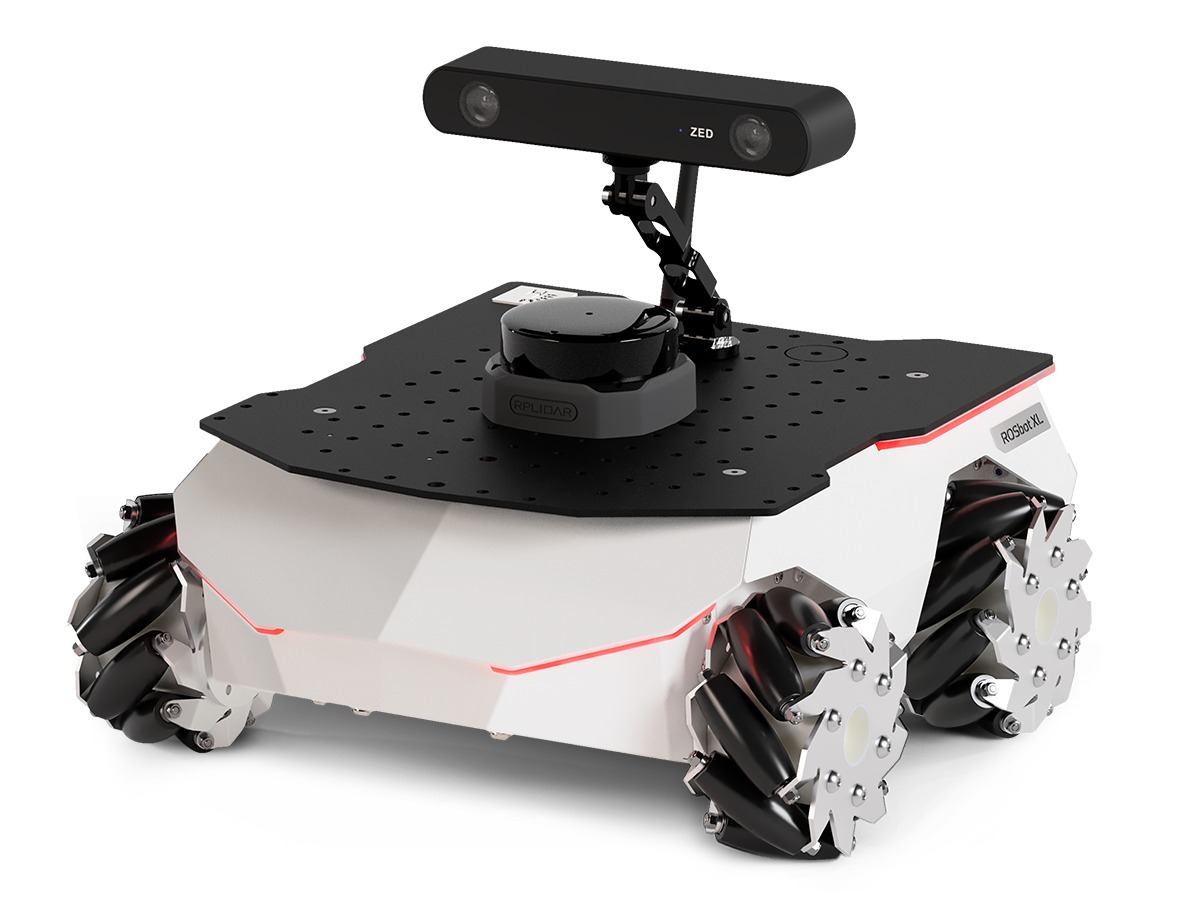}}
\caption{Turtlebot3 waffle robot was used in simulation while tests have been conducted with a turtlebot, turtlebot4 and RosbotXL in the real environment}
\label{img:maps}
\vspace{-5mm}
\end{figure}

\subsection*{Handling Dynamic Obstacles}
\label{app:obstacles}

Through equation~\ref{eq:efe_wt_c} our model can predict obstacles, and if an obstacle was not detected (e.g., not detected by the Lidar), it can still recover from a failed motion. Equation~\ref{eq:B_up} is the main factor of our map flexibility to change with the learning rates defined in Table~\ref {tab:tran_lr}.  

\begin{figure}[!htbp]
  \centering    
\subfigure[Agent map with an obstacle at position (-1,0) before moving it after a partial exploration]{\label{img:ob_21_steps}\includegraphics[width=.48\linewidth]{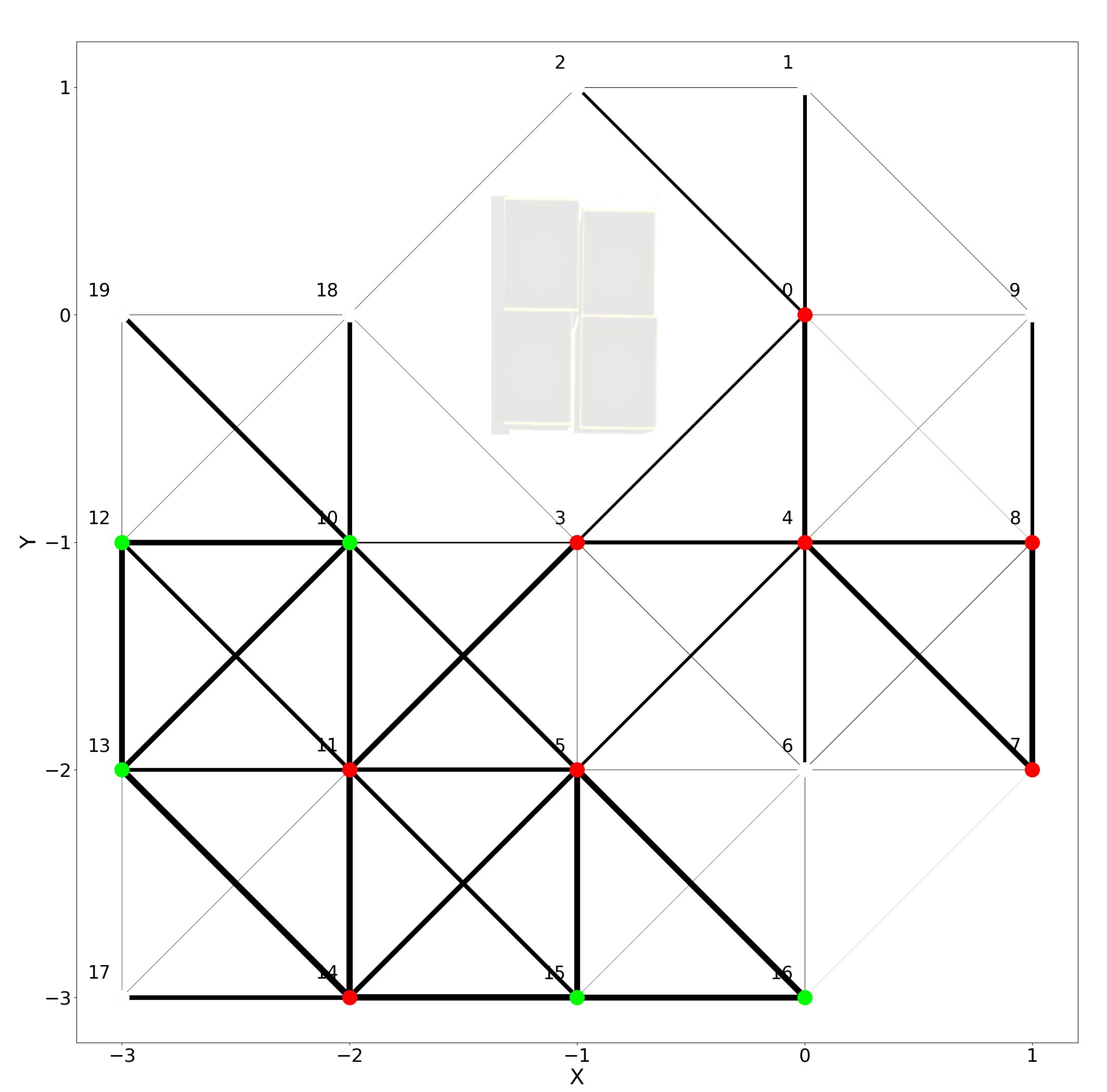}}
\subfigure[Obstacle at position (-1,-1) after 20 more steps.]{\label{img:ob_40_steps}\includegraphics[width=.48\linewidth]{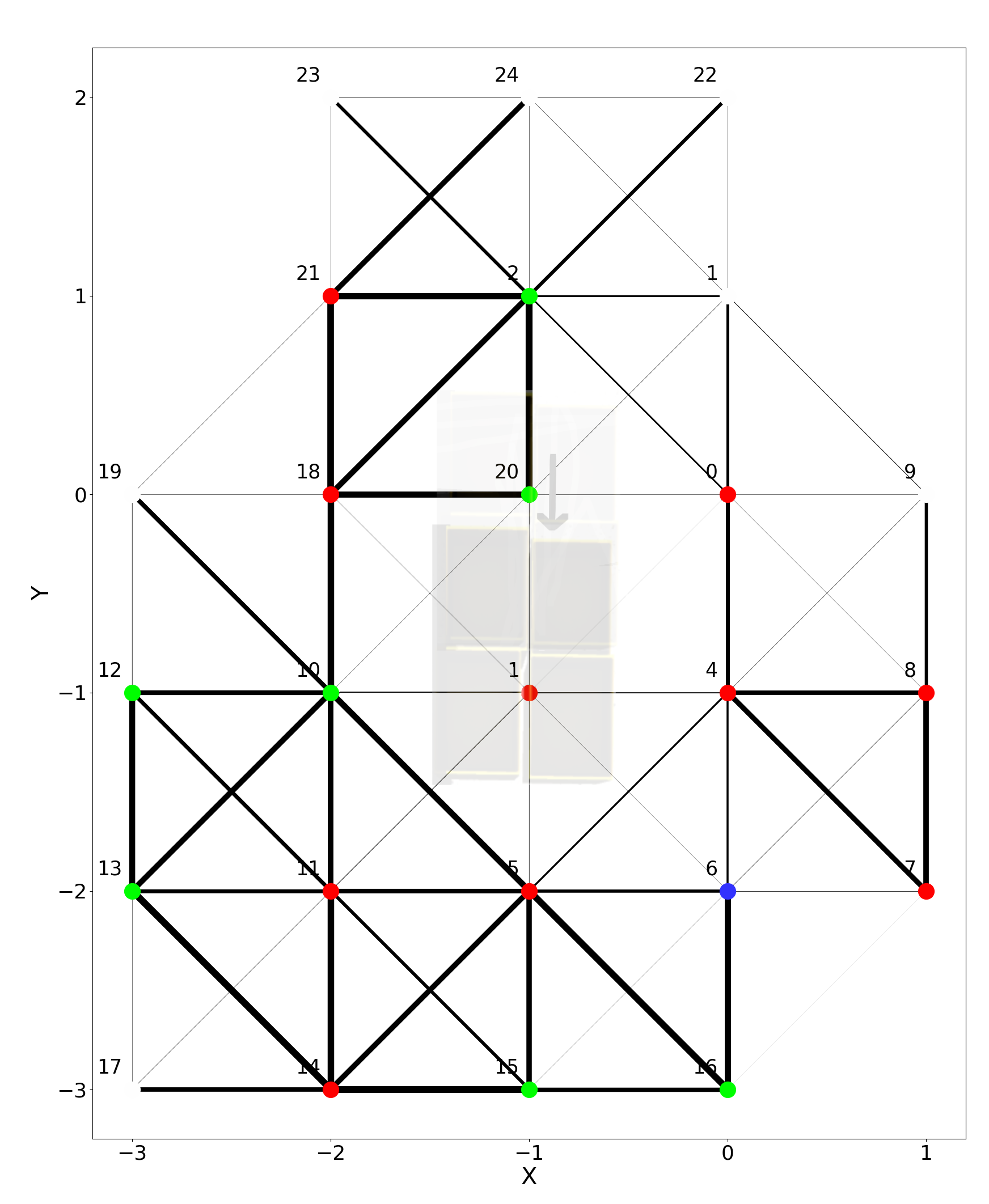}}
  \caption{An obstacle was initially placed at position (-1,0) and moved to position (-1,-1) over state 3 during exploration. The failure to reach the state reduces the transition probabilities, represented as thinner black lines between state nodes in the graph. The thickness of the link represents the transition certainty between locations, and each dot colour represents an observation; similar enough observations hold the same colours.} 
  \label{img:ob_move}
\end{figure}

\begin{figure}
    \centering
    \includegraphics[width=0.5\linewidth]{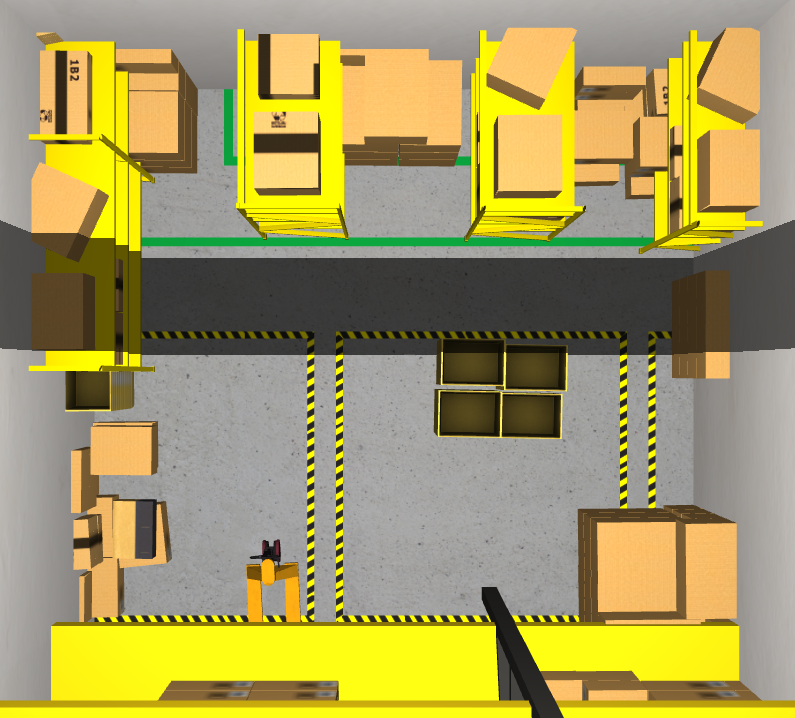}
    \caption{Top view of a mini warehouse of 36m$^2$.}
    \label{img:mini_ware}
\end{figure}

Figure~\ref{img:ob_move} exemplifies this process with an obstacle (e.g. a box) moved between two positions (from position (-1,0) to (-1,-1) over a visited state 3) while the agent explored a mini warehouse~\cite{warehouse} presented in Figure~\ref{img:mini_ware}. As the agent fails to reach state 3 after the movement of the box, it adjusts its internal map to reflect the new reality. The transition probability to that location reduces, and the state ID at this position becomes incorrect (state 1 instead of 3) because the agent cannot correct its belief by moving to that location. A new state (state 20) is created at the former position of the obstacle, and new transitions are established. This change does not affect any of the other existing states.

This display, in a qualitative way, shows how the model effectively adapts to change in the environment. The results displayed in Figure~\ref{img:ob_move} is obtained after the agent circled once around the obstacle to update all adjacent states toward the obstructed state.  

\end{document}